\documentclass{article}

\usepackage{microtype}
\usepackage{float}
\usepackage{float}
\usepackage{graphicx}
\usepackage{subcaption}
\usepackage{booktabs} 
\usepackage{tabularx}
\usepackage{graphicx}

\usepackage{hyperref}

\usepackage[accepted]{icml2024}

\usepackage{amsmath}
\usepackage{amssymb}
\usepackage{mathtools}
\usepackage{amsthm}
\usepackage{microtype}

\usepackage[capitalize,noabbrev]{cleveref}

\theoremstyle{plain}

\theoremstyle{definition}

\theoremstyle{remark}

\usepackage[textsize=tiny]{todonotes}

\usepackage{url}

\icmltitlerunning{AI Question-Answering for Access to Justice}

\begin{document}

\twocolumn[
\icmltitle{Experimenting with Legal AI Solutions: The Case of Question-Answering for Access to Justice}



\icmlsetsymbol{equal}{*}

\begin{icmlauthorlist}
\icmlauthor{Jonathan Li}{queensIGL}
\icmlauthor{Rohan V Bhambhoria}{queensIGL,queensECE}
\icmlauthor{Samuel Dahan}{queensIGL,queensLaw,cornellLaw}
\icmlauthor{Xiaodan Zhu}{queensIGL,queensECE}
\end{icmlauthorlist}

\icmlaffiliation{queensIGL}{Ingenuity Labs Research Institute, Kingston,  Canada}
\icmlaffiliation{queensECE}{Department of Electrical and Computer Engineering, Queen's University, Kingston, Canada}
\icmlaffiliation{queensLaw}{Department of Law, Queen's University, Kingston, Canada}
\icmlaffiliation{cornellLaw}{Cornell Law School, Cornell University, Ithaca, United States}
\icmlcorrespondingauthor{Xiaodan Zhu}{xiaodan.zhu@queensu.ca}

\icmlkeywords{}

\vskip 0.3in
]



\printAffiliationsAndNotice{}  

\begin{abstract}
Generative AI models, such as the GPT and Llama series, have significant potential to assist laypeople in answering legal questions. However, little prior work focuses on the data sourcing, inference, and evaluation of these models in the context of laypersons. To this end, we propose a \textit{human-centric} legal NLP pipeline, covering data sourcing, inference, and evaluation. We introduce and release a dataset, LegalQA, with real and specific legal questions spanning from employment law to criminal law, corresponding answers written by legal experts, and citations for each answer. We develop an automatic evaluation protocol for this dataset, then show that retrieval-augmented generation from only 850 citations in the train set can match or outperform internet-wide retrieval, despite containing 9 orders of magnitude less data. Finally, we propose future directions for open-sourced efforts, which fall behind closed-sourced models.
\end{abstract}

\section{Introduction}

Today, much of natural language processing rests on \textit{high-quality data}. Advances in unstructured data for pre-training have allowed general language models to be more lightweight, performant, and therefore accessible \citep{abdin2024phi3,gemmateam2024gemma}. Language models are promising in providing legal advice to laypeople, but prior work suggests that they struggle with hallucination \citep{dahan2023lawyers}. Inspired by the high-quality data that powers general domains, we identify a gap in high-quality structured legal data (e.g., question-answer pairs) approved by legal experts. In this work, we hope to build more \textit{human-centric legal AI} systems by improving the data source to address laypeople. We use this data at \textit{retrieval time} to improve model performance, which does not require additional training or fine-tuning.

Little prior work focuses on optimizing legal AI systems from start to finish for factors that matter to laypeople. Among these factors are accessibility of the services due to cost, factual correctness, and ease of understanding. In this paper, we propose an end-to-end \textit{human-centeric legal AI} framework, which covers data sourcing, training/inference, and evaluation to improve these factors; importantly, we put laypeople first by ensuring each step of the process is backed by high-quality data from legal experts (see Figure~\ref{fig:legalFramework}). To our knowledge, this type of \textit{human-centric legal framework} is the first of its kind.

First, we construct a high-quality evaluation dataset of 323 questions asked by laypeople on real legal questions and answers vetted by legal experts. We ask law students to write expert answers to these questions and release this dataset to the public. Then, we develop an automatic evaluation protocol based on the \textit{factuality} of the generated answer, as a legal expert would. Inspired by massive improvements to model quality through higher quality data at \textit{training time} (e.g., Phi-3; \citealp{abdin2024phi3}), we improve the data sourcing process at \textit{retrieval time}. Specifically, we propose domain-specific retrieval, bolstering the performance of existing LLMs on legal question-answering by retrieving from sources trusted by legal experts. We show that retrieval from under a thousand legal-expert-approved articles matches or exceeds the performance of retrieval from hundreds of millions of internet articles.

Overall, our contributions are as follows:

\begin{figure*}[t]
    \centering
    \includegraphics[width=\textwidth]{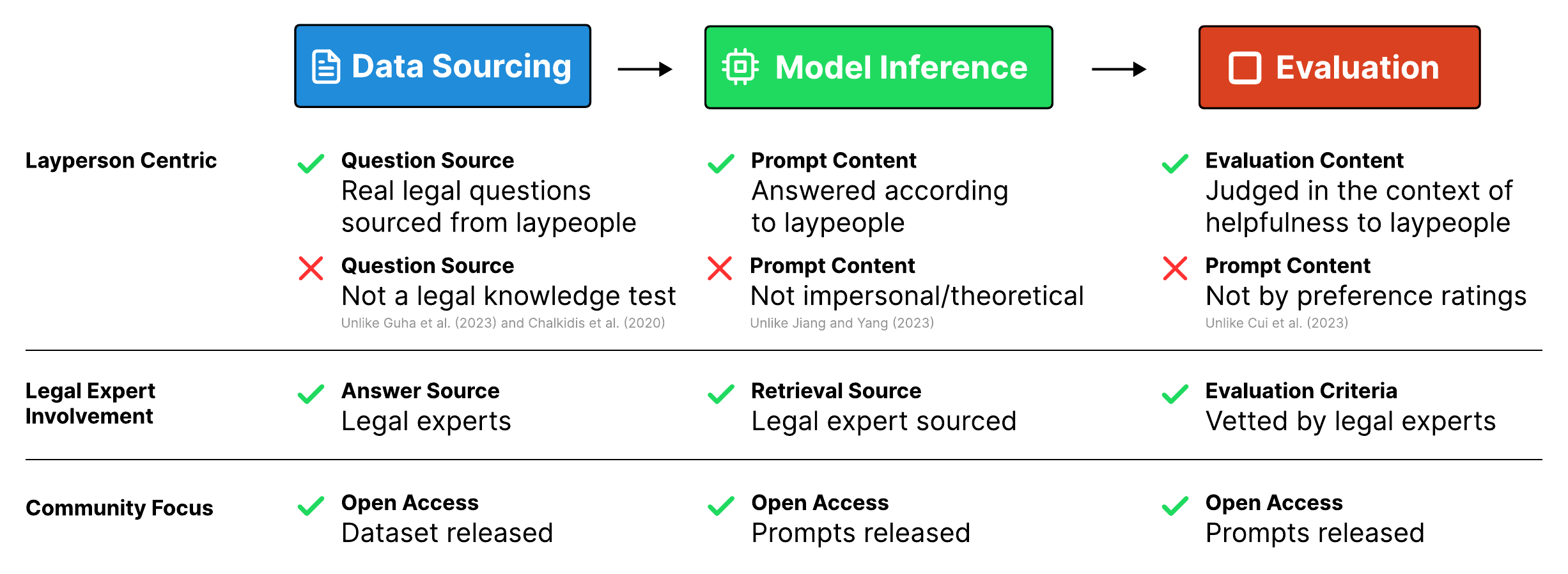}
    \caption{An overview of our framework for human-centric legal AI.}
    \label{fig:legalFramework}
\end{figure*}

\begin{itemize}
    \item We construct a dataset containing real legal questions and high-quality answers labelled by legal experts. We release the evaluation dataset publically.
    \item We create an evaluation protocol vetted by legal experts and find that existing models have room for improvement in factuality.
    \item We show that domain-specific retrieval from relatively few sources trusted by legal experts can outperform existing non-retrieval models and match or exceed retrieval-augmented models that rely on the entirety of the internet.
\end{itemize}

\section{Prior Work}

\paragraph{Retrieval Augmented Generation.}
Large language models often hallucinate and contain outdated information \citep{zhang2023siren}. Retrieval augmented generation (RAG) is an emerging approach to reduce the prevalence of hallucinations by grounding a model's generations in a data source besides the model's weights. Retrieval has been used extensively in single-hop \citep{ke2024bridging}, multi-hop \citep{sun2023recitationaugmented}, and long-form open-ended question answering \citep{lin2023ra}. With the rise of instruction-following language models \citep{touvron2023llama,chung2022scaling,brown2020language}, retrieval methods often insert context directly into the context of the language model \citep{ma-etal-2023-query,chen2023understanding}. We focus on this setting because it is possible to integrate with existing well-performing models (such as OpenAI's GPT-3.5), further supporting our human-centric goal of making legal AI more accessible.

Very recently, commercial RAG efforts such as Cohere's \texttt{Command R+} models, have been applied to legal domains with a focus on trustworthiness and data privacy \citep{cohere_law_2024}. Their retrieval method passes the retrieved documents directly into the context of a language model, such that the model generations are grounded in the context provided. In this work, we build off this line of research by focusing on retrieval from a trusted source.

\paragraph{Datasets and Legal AI Benchmarks.}
NLP has been applied to various fields in law, such as question answering, relation extraction, or text summarization \citep{zhong-etal-2020-nlp}. Previously, work was focused on domain-specific fine-tuned models \citep{chalkidis-etal-2020-legal,10.1145/3462757.3466088}. Recently, existing work has focused more on the ability of \textit{general LLMs} to perform legal reasoning \citep{yu2022legal,10.1145/3594536.3595170,blairstanek2023gpt3,yu-etal-2023-exploring}. Regarding benchmarks for LLMs in legal applications, existing benchmarks exist in Chinese \citep{Duan_2019,dai2024laiw} and American \citep{guha2023legalbench} law. However, many existing benchmarks lack evaluation of open-ended responses which are of interest to laypeople who ask language models for legal advice directly.

We source our legal questions from a public legal advice forum (though answers are written in-house by legal experts). These online forums have been used extensively as sources of data for machine learning. For instance, \citet{yao2020detection} uses data from a mental-health advice sub-community on Reddit (known as a ``subreddit'') to detect suicidality in opioid users. \citet{li-etal-2022-parameter} uses the ``r/legaladvice'' subreddit for classification of Reddit posts in evaluation. We extend this work on forum-based data by considering a much more difficult task: generating a factually accurate legal answer to a given legal question.

\paragraph{Better Data for Better AI.} Existing work has found that sourcing better data can lead to model improvements in the pretraining stage. \citet{eldan2023tinystories} create a high-quality machine-generated dataset, then shows that very small models (order of tens of millions of parameters) can learn perfect grammar from this high-quality data. Taking this a step further, \citet{li2023textbooks} and \citet{gunasekar2023textbooks} introduce the Phi series models, which gain impressive performance despite their small size, driven by high-quality data. Very recently, open-sourced state-of-the-art language models Llama 3 \citep{llama3modelcard} and Phi-3 \citep{abdin2024phi3} build off this idea, reaching state-of-the-art performance by improving data sourcing. Inspired by this work, we review a related but orthogonal direction: improving the reliability of data sourcing at \textit{retrieval time}, rather than just at \textit{pre-training time} like in prior work.

\paragraph{Automatic Evaluation.}
As LLMs become better at problem-solving, their potential to evaluate responses relative to a gold answer becomes increasingly attractive \citep{oh2024generative}. Current work has focused on evaluation of open-ended model generations for general domains, such as trivia question answering \citep{wang2023evaluating} or everyday conversational settings \citep{lin-chen-2023-llm}. In legal AI, however, most models are evaluated on closed-ended tasks that can be trivially graded \citep{info14040250,xu2023argumentative,10.1145/3594536.3595161}. \citet{Cui2023ChatLawOL} evaluates model generations using crowd-sourced human preference ratings in Chinese (with an ELO system), a step in evaluation for open-ended generations. \citet{bhambhoria2024evaluating} explores the possibility of automatic evaluation in the legal domain, showing that most classified samples align with the expert opinion. In this work, we aim to capitalize on the benefits of automatic evaluation while optimizing the process for legal factuality evaluation by consulting with legal experts.

\section{Methods}
To build a source of structured and expert-approved legal data that is effective at \textit{retrieval time}, we construct a new dataset from real legal questions and ask law professors and law students to answer these questions. Then, we use this data during our retrieval process to ground model answers in citations vetted by legal experts. During the evaluation process, we also ground our evaluations with this dataset, establishing an end-to-end legal-expert-driven framework (see Figure~\ref{fig:legalFramework}).

\begin{figure}[h]
    \centering
    \includegraphics[width=\linewidth]{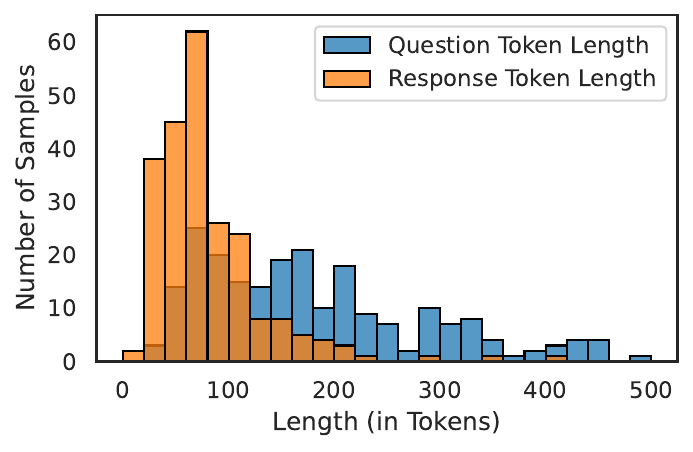}
    \caption{Distribution of question lengths and response lengths. Responses are concise and specific.}
    \label{fig:legalQaStats}
\end{figure}

Specifically, we source questions from an online community\footnote{\url{https://www.reddit.com/r/legaladvice/}}, collected from January 2021 to October 2022. These questions are specific (e.g., Table~\ref{tab:exampleData}) situations that real laypeople have, not hypotheticals\footnote{As per the legal advice community guidelines, the questions must be real questions, not hypothetical questions}. For instance, the example sample in Table~\ref{tab:exampleData} outlines a specific scenario. This real-world focus allows our evaluations to be closer to a target domain that is helpful to laypeople. Then, we ask law professors and law students to provide golden answers to these questions. Since this research was done in Canada, the legal experts we worked with were knowledgeable primarily in Canadian law. Therefore, we asked these annotators to answer these legal questions according to Canadian law. Human answers are typically concise (shorter than the questions) and under 100 tokens (see Figure~\ref{fig:legalQaStats}). Each answer contains a citation with more information relevant to the question. To perform a rigorous performance analysis across legal areas of practice, we classify each question into six categories relevant to laypeople, shown in Table~\ref{tab:categoryStructure}. The classification was done through a zero-shot classification approach \citep{laurer_building_2023} and manually inspected for correctness. To aid the community in evaluating existing LLMs, we release our evaluation dataset ($n=323$) publicly\footnote{\url{https://huggingface.co/datasets/jonathanli/law_qa_eval}}. 

\begin{table}[t]
    \centering
    \renewcommand\arraystretch{1.1}
    \caption{Composition of the area of law for each question in our dataset.}
    \begin{tabular}{l|c}
        \toprule
        Category & Percent \\
         \midrule
Employment and labour law & 27.9 \\
Family and juvenile law & 27.1 \\
Real estate law & 21.4 \\
Corporate law & 9.2 \\
Personal injury law & 9.2 \\
Civil rights law & 5.2 \\
         \bottomrule
    \end{tabular}
    \label{tab:categoryStructure}
\end{table}

\begin{table}[!t]
\centering
\small
\caption{Example question (``source'') and provided answers and citation.}
\vspace{9pt}
\begin{tabularx}{\linewidth}{lX}
\toprule
Source & 
Father took family pet during divorce and probably will give him away.
My parents recently got a divorce a few months back and my father has been sending very disgusting harassing messages any way he can. He already has escapee warrants and now he is also getting more added to him due to his messages. He took the family dog who is legally under my mothers name and we know he is not safe with him. We have no idea where he is at but we know he is not a very safe person with our dog. We really fear of him giving the dog away to his friend and we wouldn't see him ever again. What do we do if that's the case? Can we fight to get him back if he gives him to his friend?
\\
\midrule
Answer & In Ontario, dogs are considered personal property. In determining which spouse has a right to the dog, a court will consider ownership papers as well as several factors such as: Is the pet more bonded to one person over the other? Who can best provide continued care? Who paid for the pet? \\
\midrule
Citation & \texttt{https://www.siskinds.com/pet-c ustody-laws-in-ontario} \\
\bottomrule
\end{tabularx}
\label{tab:exampleData}
\end{table}

\begin{figure*}[t]
    \centering
     \begin{subfigure}[b]{\textwidth}
       \centering
    \includegraphics[width=\textwidth]{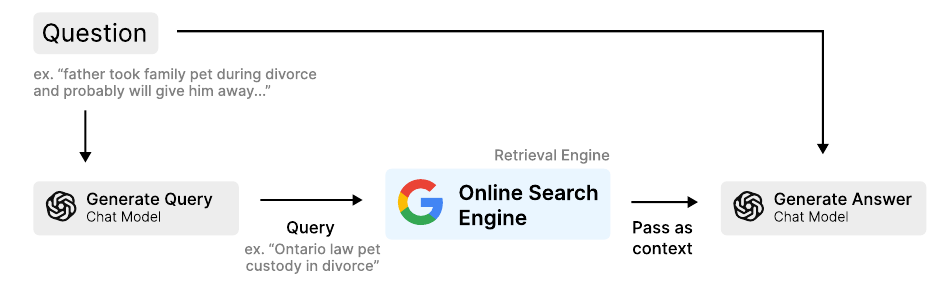}
    \caption{Internet-wide retrieval}
    \label{fig:internetRetreivalFramework}
    \end{subfigure}
     \begin{subfigure}[b]{\textwidth}
       \centering
        \includegraphics[width=\textwidth]{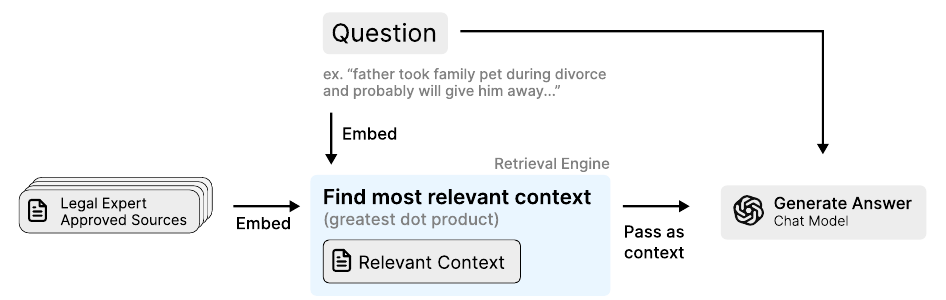}
    \caption{\textit{Human-centric} legal retrieval \textbf{(ours)}}
    \label{fig:legalRetreivalFramework}
    \end{subfigure}
    \caption{Retrieval-based methods used for our experiments. Given a legal question, retrieval is performed to generate a relevant answer.}
\end{figure*}


Inspired by \citet{lin-chen-2023-llm}, we employ human-grounded automatic evaluation of the generated answers. We use golden labels written by legal experts as "grounding" for the language model, then ask the LLM (\texttt{GPT-4-0613}) to rate the answer's factuality relative to the expert answer. If there is a factual contradiction with the expert answer, we treat the model response as factually disagreeing. We run the automatic evaluator with a temperature of zero (since we only need the prediction "factual" or "not factual"). In our study, we minimize this factual disagreement to build factually accurate models. This approach is among the criteria that legal experts use to evaluate answers by law students \citep{bhambhoria2024evaluating}.

%

\paragraph{Retrieval Methods.} In addition to using existing language models to answer legal questions, we also consider grounding a model's responses in the context of a relevant article. As previously discussed, we believe the retrieval process could also benefit from high-quality legal data. In this work, we try to retrieve from only trusted legal sources, as shown in Figure~\ref{fig:legalRetreivalFramework}. Since we split the dataset into a train and test set, we use the citations from the train set ($n=850$) as a set of trusted legal documents and only retrieve from these documents. This offers two main benefits: (a) the documents used by the model are known to be factual and helpful by legal experts, which is not the case for any document on the internet, and (b) searching a smaller subset of legal-expert-approved documents provides computational and storage benefits.

As shown in Figure~\ref{fig:legalRetreivalFramework}, we embed both the context and the question using an existing state-of-the-art embedding model, \texttt{BAAI/bge-large-en-v1.5} \citep{bge_embedding}. Then, we compute the dot product between the question and each document, selecting the document with the greatest dot product as the most relevant sample. Then we provide this document in the context of an existing language model (\texttt{GPT-3.5-turbo}), using prompts containing both the context and question. Unlike prior work, we evaluate retrieval from only legal expert sources rather than the entirety of the internet. This constitutes our model inference part in Figure~\ref{fig:legalFramework}. We call this "legal retrieval".

As a baseline, we evaluate (a) GPT-3.5-turbo without retrieval augmented generation and (b) GPT-3.5-turbo with retrieval from the entire internet. For (b), we use a language model to produce a query for a legal question that can be queried for in a search engine (Figure~\ref{fig:internetRetreivalFramework}). Then we use an existing web search engine (Google) to find the most relevant article and inject it into the context of the language model while answering this question. This article is used to provide context to the model. We call (b) "internet retrieval".

\begin{figure*}[t]
    \centering
    \includegraphics[width=\textwidth]{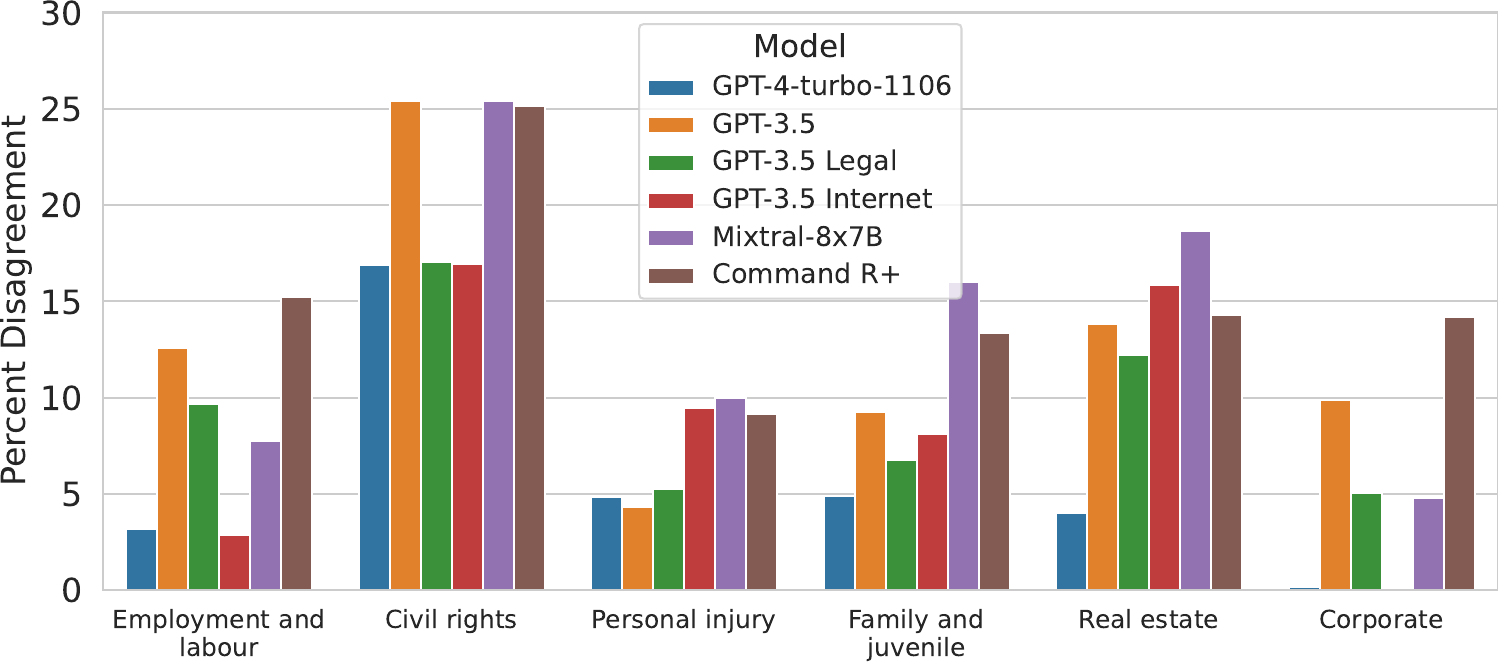}
    \caption{Factual disagreement of each model by category. Lower is better.}
    \label{fig:categoryLegalqa}
\end{figure*}

Internet retrieval is not as simple as retrieval from only legal documents, since a Google search is performed. Search engines likely contain more sophisticated methods than a simple embedding similarity check. When evaluating the performance of retrieval from the entirety of the internet against retrieval from only legal documents, more computation occurs using an internet-wide search.

\paragraph{Generative Baselines.} As a further baseline, we evaluate the state-of-the-art open-source models on this legal task. We use the state-of-the-art \texttt{Mixtral-8x7B} \citep{jiang2024mixtral} for non-retrieval generation and the state-of-the-art Cohere Command R+ model \citep{aiden_2024} for retrieval-augmented generation. We rely on Cohere's pipeline for retrieval-augmented generation, whose model is purposefully tuned for this purpose.

For each method, we pass three samples from a separate train split as fewshot examples into the input prompt. We use the default generation and sampling settings (i.e., temperature and top-p) for each model. 

\section{Results and Discussion}
\label{sec:results}

\begin{figure}[H]
    \centering
    \includegraphics[width=\linewidth]{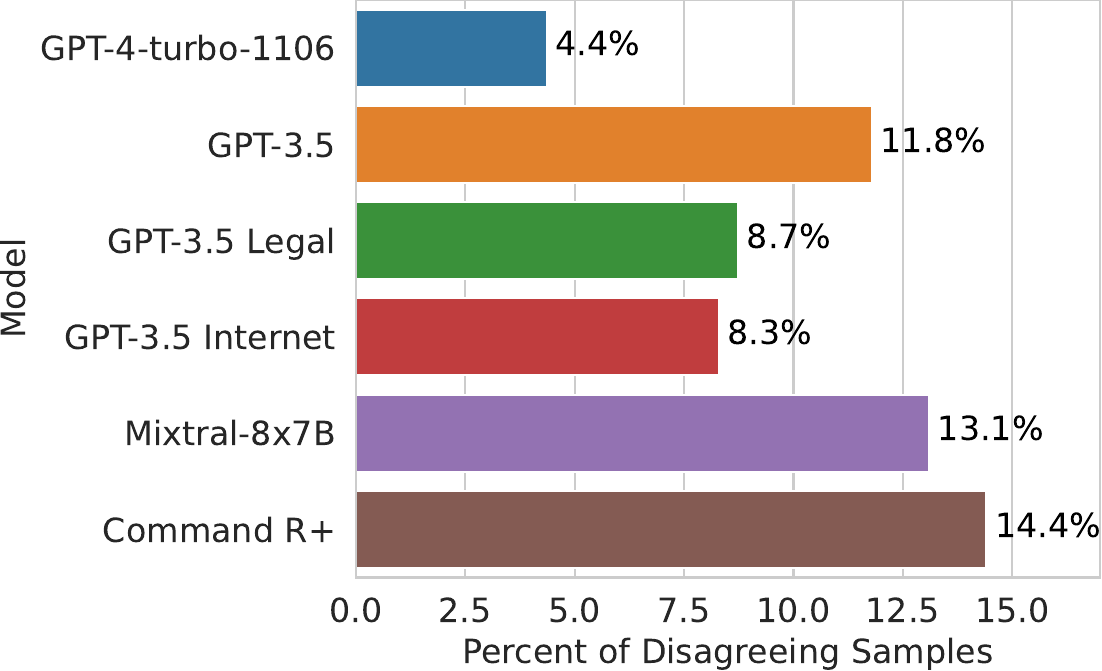}
    \caption{Factual disagreement for each model. ``GPT-3.5 Legal'' is retrieval using only legal documents, and ``GPT-3.5 Internet'' is retrieval from the entire internet.}
    \label{fig:legalQaResults}
\end{figure}

As seen in Figure~\ref{fig:legalQaResults}, the retrieval-based approaches tested typically perform better than their tested non-retrieval counterparts. Additionally, we make various observations:

\paragraph{Open-source models fall behind.} Mixtral-8x7B and Command R+, state-of-the-art open language models perform worse than current closed-source models. Future work should continue testing the newest state-of-the-art open-sourced models for this task. Though we recognize the importance of open-source models, we also note that practically, the inference costs associated with using open-sourced models are nonzero for a layperson. Therefore, closed-sourced models like GPT-3.5 are still appealing from a cost perspective compared to other open-sourced models. 

\begin{figure*}[t]
    \centering
    \includegraphics[width=\textwidth]{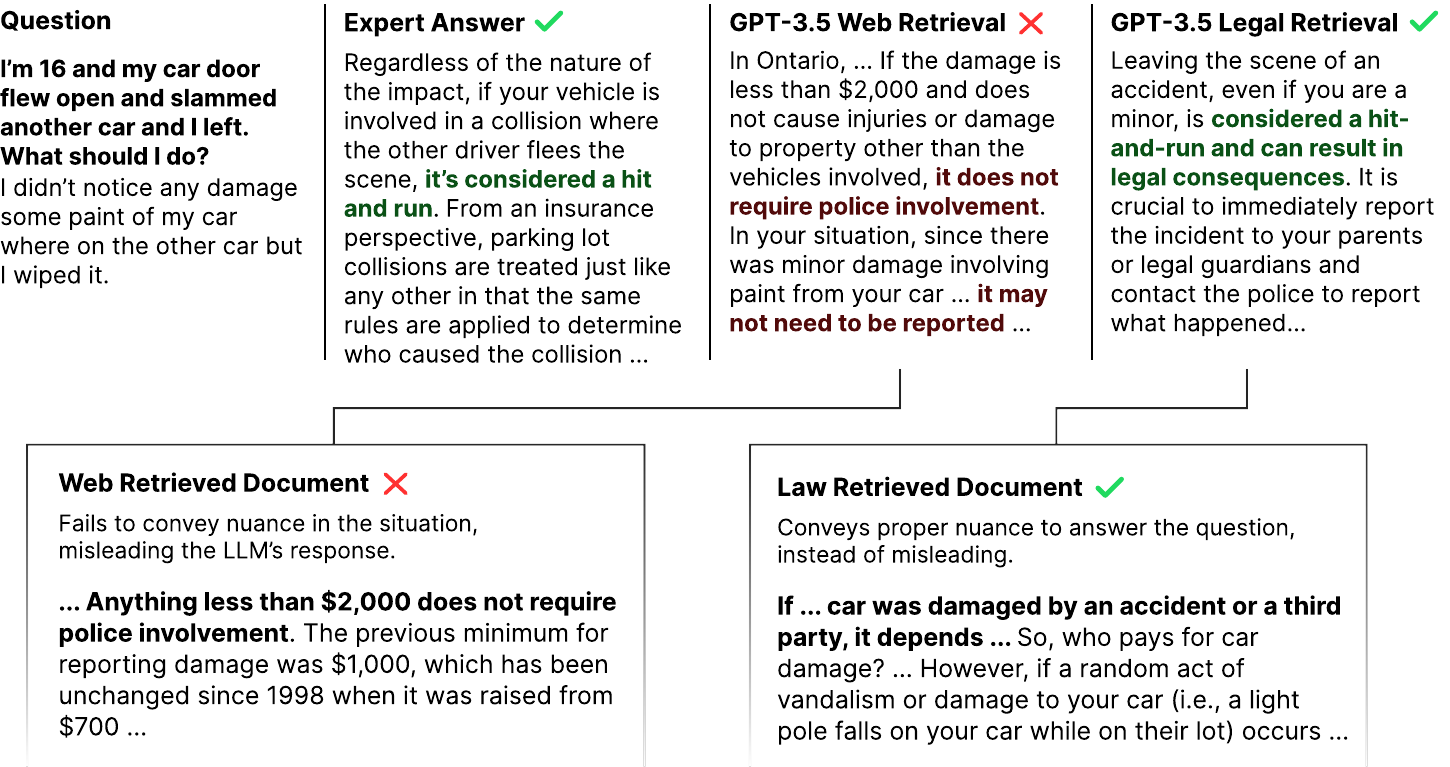}
    \caption{Qualitative example showcasing retrieval from the entire internet and our high-quality source of legal documents. In this case, retrieval from the entire internet provides a less nuanced source of information.}
    \label{fig:qualitativeExample}
\end{figure*}

\paragraph{Using limited legal documents is just as useful as the entire internet.} Retrieval from limited legal sources (just 850 documents) performs similarly to searching the entirety of the internet (hundreds of billions of documents). In the case of Cohere Command R+, a state-of-the-art enterprise retrieval system used in law \footnote{\url{https://txt.cohere.com/how-llms-can-boost-legal-productivity-with-accuracy-and-privacy/}}, our simple retrieval method improves performance, despite Command R+ performing retrieval across the entire internet. The indexable internet contains more information than just our limited legal information, but the vastness of the space also make retrieval of documents much more difficult. Surprisingly, narrowing down the number of retrievable documents by 9 orders of magnitude\footnote{This was roughly calculated based on the ``hundreds of billions'' of pages that Google indexes \citep{Google_2023} compared to our dataset of under a thousand samples.} retains enough information to perform similarly to retrieval from the entire internet. By reducing the number of documents that can possibly be retrieved, human-driven review of individual documents is more feasible, and storage and computational costs decrease.

\paragraph{GPT-4 outperforms retrieval.} We find that \texttt{gpt-4-turbo-1106} outperforms even retrieval models. However, we note that use of GPT-4 often has prohibitive costs (in some cases, 60x more), especially when paired with retrieval-augmented generations which span multiple contexts. Therefore, we still believe that relative improvements to GPT-3.5-turbo, a much more cost-effective model, continue to benefit the layperson despite the existence of more expensive and well-performing models.

\paragraph{Some categories of questions are more difficult to answer accurately.} We find that some categories, such as ``civil rights'' and ``real estate'' are the most challenging for existing language models, illustrated in Figure~\ref{fig:categoryLegalqa}. Qualitatively, we observed that questions falling under these areas of law contained the most specific and personal questions, and also had more nuanced cases (e.g., a highly specific question about a tenant's landlord).

\paragraph{In what situations does legal retrieval help?} Apparent in Figure~\ref{fig:qualitativeExample}, in some cases retrieval from the entire internet can provide a lack of nuanced information, making the response factually incorrect compared to retrieval from only vetted legal documents. In the specific example presented, the retrieved web article failed to capture nuance when describing the punishment for damaging a vehicle during a traffic infraction, while the retrieved legal article from a vetted legal source did.

Figure~\ref{fig:categoryLegalqa} shows the strengths and weaknesses of each approach by category. Using retrieval from our expert source of legal documents is almost always better than or similar to the performance of non-retrieval methods. Additionally, retrieval from vetted legal sources performs on par with or better than retrieval from the entire internet, except for ``employment and labour'' and ``corporate'' categories. Further experiments are required to investigate the cause of this disparity, though we hypothesize it is because these categories have questions that span a larger space of questions, implying that retrieval from the internet is more applicable.

\subsection{Future Directions}
\paragraph{Closing the gap between open-sourced and closed-sourced models.} As shown in Section~\ref{sec:results}, the gap between the top open-sourced model and the closed-sourced models (\texttt{GPT-4}) is substantial. From the perspective of human-centric legal AI, this is problematic as these black-box models often lack accountability in their data sources, which is especially important in the legal domain \citep{dahan2023lawyers}.

\paragraph{Continual Updating.} Laws and regulations constantly evolve, and legal AI systems need to stay up-to-date with the latest changes. Investigating techniques that can efficiently integrate new legal information and adapt models accordingly would better reflect the dynamic nature of the legal landscape. Currently, our methodology focuses on retrieving from a static set of documents, though the set of documents could be continually pruned and updated \citep{bhambhoria2024evaluating}.

\paragraph{Unstructured Legal Data.} In this work, we focus on sourcing high-quality structured data for use during retrieval and evaluation. Expert involvement in the data selection process could extend to unstructured data during the pre-training phase for the legal domain, which has already shown promise in general domains \citep{gunasekar2023textbooks}.

\subsection{Conclusions}
Towards the goal of building more accessible and human-centric legal AI, a high-quality novel dataset containing question-answer pairs was produced and publically released, addressing a previous lack of expert-involved structured data. Existing open-sourced and closed-sourced language models were evaluated using an automatic evaluation framework based on the factuality of answers. We found that retrieval from a small set of legal documents can match or outperform the performance of retrieval from the entire internet, despite requiring many orders of magnitude less data.

\bibliography{example_paper,anthology}
\bibliographystyle{icml2024}

\newpage

\end{document}